\documentclass{article} 
\usepackage{iclr2025_conference,times}

\usepackage{amsmath,amsfonts,bm}

\def\eqref#1{equation~\ref{#1}}

\def\1{\bm{1}}

\DeclareMathAlphabet{\mathsfit}{\encodingdefault}{\sfdefault}{m}{sl}
\SetMathAlphabet{\mathsfit}{bold}{\encodingdefault}{\sfdefault}{bx}{n}

\usepackage{hyperref}
\usepackage{url}

\newcommand{\methodname}{\textsc{InstantIR}\xspace}

\usepackage{booktabs}
\usepackage{enumitem}
\usepackage{graphicx}
\usepackage{subcaption}
\usepackage{multirow}
\usepackage{xspace}
\usepackage{wrapfig}
\usepackage{algorithm}
\usepackage{algpseudocode}
\renewcommand{\Require}{\item[\textbf{Input:}]} 
\renewcommand{\Ensure}{\item[\textbf{Output:}]} 

\usepackage{pifont}
\newcommand{\cmark}{\ding{51}}
\newcommand{\xmark}{\ding{55}}

\renewcommand{\cite}{\citep}

\title{\methodname: Blind Image Restoration with \\ Instant Generative Reference}

\author{Jen-Yuan Huang\textsuperscript{$1,2$}\quad Haofan Wang\textsuperscript{$2$}\quad\quad Qixun Wang\textsuperscript{$2$}\quad\quad Xu Bai\textsuperscript{$2,3$}\\
\bf Hao Ai\textsuperscript{$2$}\quad\quad\quad\quad\quad\quad Peng Xing\textsuperscript{$2$}\quad\quad\quad\ \ Jen-Tse Huang\textsuperscript{$4$}\\
\textsuperscript{$1$}Peking University\quad\ \ \textsuperscript{$2$}InstantX Team\quad\quad \textsuperscript{$3$}Xiaohongshu Inc\quad\quad\quad \\ \textsuperscript{$4$}The Chinese University of Hong Kong
}

\iclrfinalcopy

\begin{document}

\maketitle

\begin{figure}[h]
\centering
    \includegraphics[width=\linewidth]{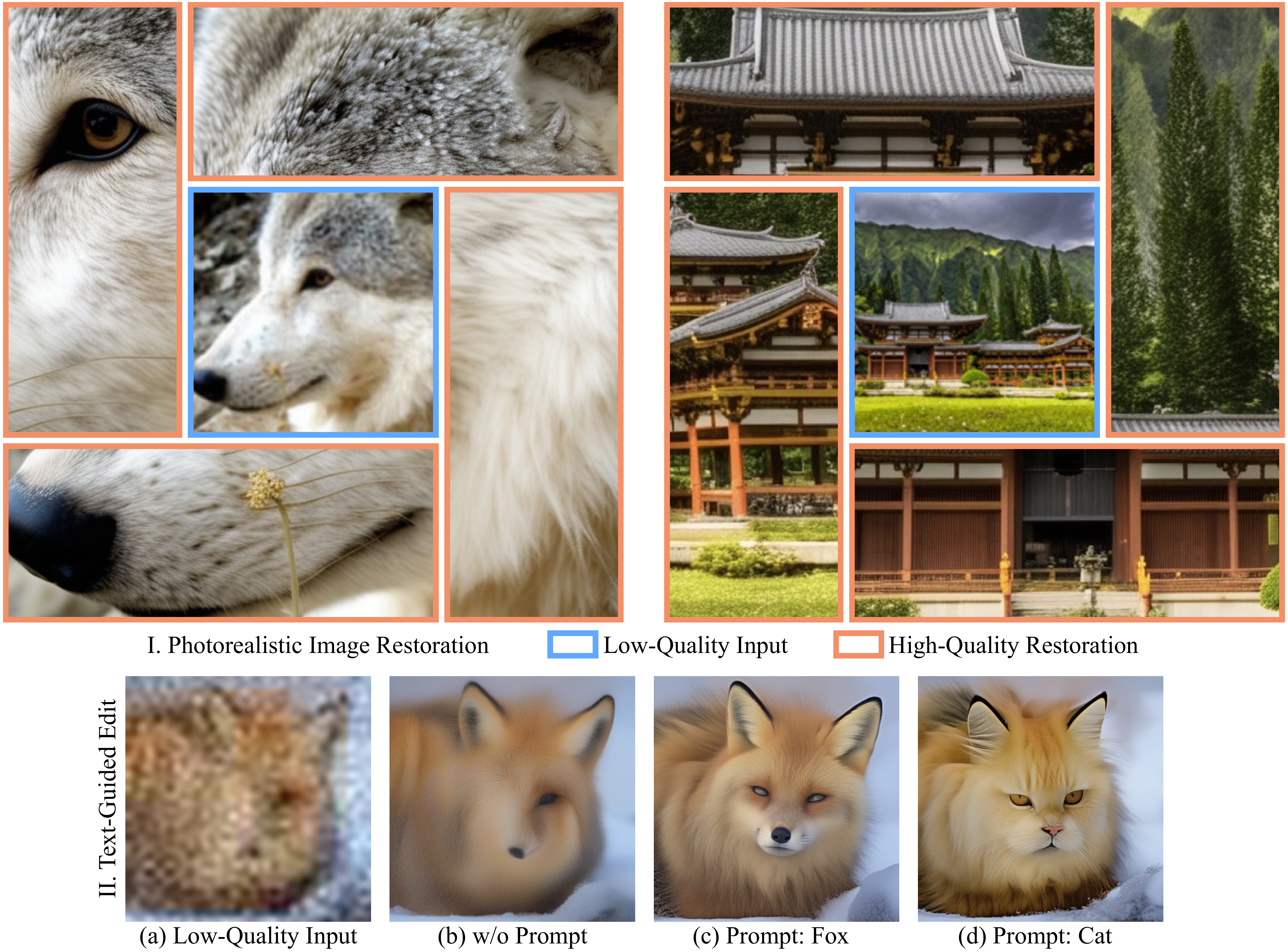}
    \caption{I. \methodname presents exceptional capability in reproducing photorealistic details. II. \methodname provides an active interface for natural language guidance, helps handling large degradation and features creative restoration with semantic editing.}
\end{figure}

\begin{abstract}
Handling test-time unknown degradation is the major challenge in Blind Image Restoration (BIR), necessitating high model generalization. An effective strategy is to incorporate prior knowledge, either from human input or generative model. In this paper, we introduce Instant-reference Image Restoration (\methodname), a novel diffusion-based BIR method which dynamically adjusts generation condition during inference. We first extract a compact representation of the input via a pre-trained vision encoder. At each generation step, this representation is used to decode current diffusion latent and instantiate it in the generative prior. The degraded image is then encoded with this reference, providing robust generation condition. We observe the variance of generative references fluctuate with degradation intensity, which we further leverage as an indicator for developing a sampling algorithm adaptive to input quality. Extensive experiments demonstrate \methodname achieves state-of-the-art performance and offering outstanding visual quality. Through modulating generative references with textual description, \methodname can restore extreme degradation and additionally feature creative restoration.
\end{abstract}

\section{Introduction}

Image restoration seeks to recover High-Quality (HQ) visual details from Low-Quality (LQ) images. This technology has a wide range of important applications. It can enhance social media contents to improve user experience~\cite{chao2023equivalent}. It also functions at the heart in industries like autonomous driving~\cite{patil2023multi} and robotics~\cite{porav2019can} by improving adaptability in diverse environments, as well as assists object detector in adverse conditions~\cite{sun2022rethinking}.

Image restoration remains a long-standing challenge extending beyond its practical application. The information loss during degradation makes a single LQ image corresponds to multiple plausible restorations. This ill-posed problem is further exacerbated in Blind Image Restoration (BIR), where models are tested under unknown degradation. A common strategy is to leverage prior knowledge. Reference-IR models use other HQ images to modulate LQ features, requiring additional inputs with similar contents but richer visual details~\cite{lu2021masa}. Generative approaches, on the other hand, directly learn the HQ image distribution. The input is first encoded into a hidden variables $z$, which servers as the generation condition to sample HQ image from the learned distribution $p(y|z)$. Although generative methods achieve single-image restoration, they are prone to hallucinations that produce artifacts in restoration~\cite{yang2020learning}. This happens when the encoder fails to retrieve accurate hidden variable due to the input distribution shift in degradation. Existing methods improve robustness by training on more diverse synthetic degradation data or introduce discrete feature codebook. We argue that these are only shot-term solutions. Alternative methods are pendding to be explored to better address unknown inputs in BIR.

In this paper, we present \methodname, a dynamic restoration pipeline that iteratively refines generation condition using a pre-trained Diffusion Probabilistic Model (DPM). \methodname employs two complementary way for processing input LQ image. First, a pre-trained vision encoder extracts compact representation from degraded content. The encoder's high compression rate enhances the robustness in the extracted representation, while retaining only high-level semantics and structural information. Next, we introduce the \textit{Previewer} module, a distilled DPM capable of one-step generation. At each generation step, the previewer decodes current diffusion latent using the compact representation, providing a restoration preview resembles original input in high-level features. This preview serves as an instant generative reference to guide the \textit{Aggregator} in encoding identity and other fine-grained missing from the compact representation. We observe in experiments that the previewer tends to decode aggressively when the input is clear, resulting in high variance in restoration previews. We take this as a reliable indicator of input image quality, and develop an adaptive sampling algorithm that amplifies the fine-grained encoding with relatively high quality inputs. Additionally, we find the previewer is controllable through text prompts, which produces diverse generative references and enables semantic editing with restoration.
Our contributions are as follows:
\begin{enumerate}[leftmargin=15pt]
    \item We explore a novel BIR method that iteratively aligns with the generative prior to address unknown degradation;
    \item We introduce a novel architecture based on pre-trained DPM, which dynamically adjusts the generation condition by previewing intermediate outputs;
    \item We develop sampling algorithms tailored for our pipeline, enabling both adaptive and controllable restoration to text prompts;
    \item We perform extensive evaluations to validate the effectiveness of the proposed methods.
\end{enumerate}
\section{Related Work}

\subsection{Diffusion Model}

DPM is a class of generative model that generate data by iteratively denoising from Gaussian noise~\cite{sohl2015deep, ho2020denoising, song2020score}. Typically, a neural network with a UNet architecture~\cite{ronneberger2015u} is trained to predict the noise added at each inference step. DPM offers superior mode coverage compared to Variational Autoencoders (VAE)~\cite{kingma2013auto} and outperform GAN-based models~\cite{goodfellow2020generative} in generation quality without the need of adversarial training~\cite{dhariwal2021diffusion}. These advantages establish DPM as the leading approach in vision generative models. By incorporating additional inputs, DPMs can learn diverse conditional distributions~\cite{nichol2021improved}, with the most widely used application being text-to-image (T2I) generation~\cite{rombach2022high, saharia2022photorealistic, ramesh2022hierarchical}. Leveraging the flexibility of text inputs and the vast amount of text-image training data~\cite{schuhmann2022laion}, these models are capable of generating images with exceptional visual quality and remarkable diversity, forming the foundation for many subsequent excellent work in vision generative models~\cite{wang2024instantid, wang2024instantstyle}.

\subsection{Blind Image Restoration}

The task setting makes BIR particular valuable in real-world applications. The major challenge in BIR is the input distribution gap between training and testing data. Previous work have explored multiple ways to address this issue. Feature quantification is widely used in generative-based methods~\cite{esser2021taming, van2017neural, zhou2022towards}. They align the encoded LQ image features to a learnable feature codebook, ensuring the input to generator is unaffected by domain shifts. However, this hard alignment constraints the generation diversity and quality by the capacity of the discrete codebook. Previous work have also explored the application of powerful DPM in BIR. Some approaches design specialized architectures and train DPMs from scratch~\cite{saharia2022image, sahak2023denoising, li2022srdiff}, while the others apply additional modules on pre-trained T2I model~\cite{wang2024exploiting, yu2024scaling, sun2024coser}, leveraging their large-scale prior. In many practical scenarios, HQ images with similar contents, such as those from photo albums or video frames, are available. This has spurred interest in restoring images using reference-based methods~\cite{cao2022reference,jiang2021robust,lu2021masa,xia2022coarse,yang2020learning,zhang2019image}. They adopt regression models to learn how to transfer high-quality features to LQ images, enhancing details restoration.
\section{Methodology}

\begin{figure}[t]
\centering
    \includegraphics[width=0.9\linewidth]{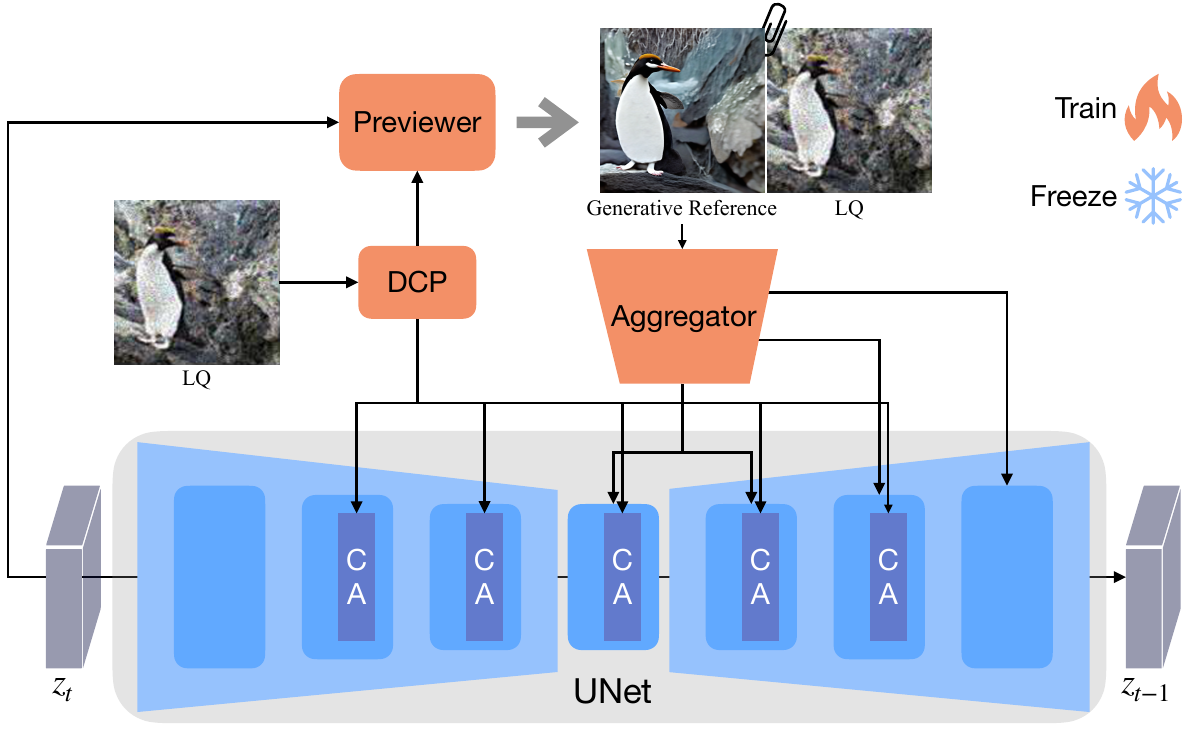}
    \caption{The overall pipeline of \methodname. \methodname adopts a novel previewing mechanism to actively align with generative prior. This is achieved by three key modules: 1) DCP for compact LQ image representation encoding; 2) Previewer for decoding it into generative prior; and 3) Aggregator for integrating the generative reference and LQ input into sampling conditions.}
\end{figure}

The distribution gap between training and testing data exacerbates the ill-posed nature of BIR, causing hallucinations in generation-based IR models and producing artifacts. We attribute this to the error in encoding LQ image, and propose a generative restoration pipeline that refines the LQ encodings with generative references. This is achieved by exploiting the reverse process of DPM. Specifically, we first encode the LQ image into a compact representation via pre-trained vision encoder, capturing global structure and semantics to initiate diffusion generation. Conditioned on this embedding, our Previewer module generates a restoration preview at each diffusion time-step. The preview resembles to the input image with more plausible details, and they are further fused in the Aggregator module to preserve fidelity. Finally, the adjusted LQ encoding is used to control the pre-trained DPM for a fine-grained diffusion step.

\subsection{Preliminaries}

DPM involves two stochastic processes named forward and reverse process~\cite{ho2020denoising}. In the forward process, \textit{i.i.d.}~Gaussian noise is progressively added to the image $\boldsymbol{x}$. The marginal distribution of diffusion latent $\boldsymbol{x}_t$ follows $\mathcal{N}\left( \alpha_{t}\boldsymbol{x}, \beta_{t}\boldsymbol{I} \right)$, where $\alpha_t$ and $\beta_t$ are hyperparameters defining the forward process. $\boldsymbol{x}_t$ converges to pure noise as $t$ increases, and the reverse process generates images by inverting the forward process. Generally, we train a neural-network to predict the noise added at each time-step by minimizing the diffusion loss:
\begin{equation}
\label{diffusion_loss}
\mathcal{L}_{diff}=\mathbb{E}\left[\lVert \boldsymbol{\epsilon}_{\theta}\left(\boldsymbol{x}_t,t\right)-\boldsymbol{\epsilon} \rVert^2\right],
\end{equation}
where $\boldsymbol{\epsilon}_\theta$ denotes the noise-prediction network. At each step in the reverse process, we can retrieve a denoising sample with the predicted noise and re-parameterization~\cite{karras2022elucidating}:
\begin{equation}
\label{ddim}
\boldsymbol{\hat{x}}=\frac{\boldsymbol{x}_t-\beta_t\boldsymbol{\epsilon}_{\theta}\left(\boldsymbol{x}_t,t\right)}{\alpha_t}.
\end{equation}
In the open-sourced T2I model Stable Diffusion (SD)~\cite{rombach2022high}, the noise-prediction network $\boldsymbol{\epsilon}_\theta$ is additionally conditioned on a text input that describes the target image. Moreover, SD employs a VAE to move the input $\boldsymbol{x}_{t}$ into latent space $\boldsymbol{z}_{t}$, compressing inputs by a factor of 48 and significantly reduces the memory usage to enable image generation up to $512^2$ resolution.


\subsection{Architecture}

The restoration pipeline of \methodname consists of three key modules: Degradation Content Perceptor (DCP) for compact LQ image encoding, Instant Restoration Previewer for generating references on-the-fly during the reverse process, and Latent Aggregator for integrating restoration references.

\paragraph{Degradation Content Perceptor}

We employ the pre-trained DINO~\cite{oquab2023dinov2} for providing compact LQ image representation. Compared to CLIP~\cite{radford2021learning}, a common choice in image editing~\cite{ye2023ip}, DINO's self-supervised training with data augmentation improves robustness of the encoded features. The extracted LQ representation is modulated by a learnable Resampler~\cite{han2024emma} and projected as context to the cross-attention layers of diffusion UNet. For the $l$-th cross-attention block, we introduce an additional cross-attention operation:
\begin{equation}
\label{lq-attn}
\boldsymbol{f}_{out}^{l} = \boldsymbol{f}_{in}^{l} + \texttt{CrossAttn}\left(\boldsymbol{f}_{in}^{l}, \boldsymbol{c}_{txt}\right)+ w^{l}\cdot \texttt{CrossAttn}\left(\boldsymbol{f}_{in}^{l}, \Phi\left(\boldsymbol{c}_{lq},t \right)\right),
\end{equation}
where $\Phi$ denotes the DCP module and $\boldsymbol{c}_{lq}$ is the LQ context matrix. We retain the text cross-attention here as it is a crucial part of the pre-trained T2I model that synthesizes high-level semantics. Jointly training DCP with textual transformation allows it to focus on low-level information absent in the other modality. We introduce a hyper-parameter $w^l$ to regulate their behaviors. Note that the DCP also takes time-step $t$ as input to establish temporal dependency in the output. Specifically, we use adaptive layer-normalization to modulate the context matrix from the DCP according to time-step $t$:
\begin{equation}
    \Phi\left(\boldsymbol{x},t \right)=\boldsymbol{\mathcal{T}}_{scale} \odot \texttt{LayerNorm}\left( \boldsymbol{c}_{lq}\right) + \boldsymbol{\mathcal{T}}_{shift},
\end{equation}
where, $\boldsymbol{\mathcal{T}}_{scale}, \boldsymbol{\mathcal{T}}_{shift}$ are calculated from the time-step. We train the DCP module on a frozen diffusion model using the standard diffusion loss in Eq.~\ref{diffusion_loss}.

\paragraph{Instant Restoration Previewer}
\label{previewer}

The compact representation encoded by the DCP, while robust against degradation, losses high-level information.  We introduce Previewer, a diffusion model generates from current diffusion latent instead of noise, to decode generative references from the DCP encoding. Decoding at each diffusion time-step requires $\left(T\left(T+1\right)/2\right)$ network forward passes with the vanilla T2I model. To streamline this process, we fine-tune the Previewer using consistency distillation~\cite{luo2023latent} to make it a one-step generator. For diffusion latent $\boldsymbol{z}_{s}$ at time-step $s$, we first obtain the Previewer output conditioned solely on $\boldsymbol{c}_{lq}$. Then, we perform a diffusion step using the pre-trained model from $\boldsymbol{z}_{s}$, conditioned on both $\boldsymbol{c}_{lq}$ and $\boldsymbol{c}_{txt}$, to reach $\boldsymbol{z}_{t}$. $\boldsymbol{z}_{t}$ is regarded as the ground-truth diffusion latent at time-step $t$ in the sampling trajectory. Finally, we get the preview of $\boldsymbol{z}_{t}$, again conditioned solely on $\boldsymbol{c}_{lq}$. The consistency distillation loss is then calculated by:
\begin{equation}
\label{preview_loss}
\mathcal{L}_{dist}=\lVert \Psi \left(\boldsymbol{z}_{s},s, \Phi\left(\boldsymbol{c}_{lq}, s\right)\right)- \texttt{StopGrad}\left( \Psi\left( \boldsymbol{z}_{t},t,\Phi\left(\boldsymbol{c}_{lq},t \right) \right) \right) \rVert^2,
\end{equation}
where $\Psi$ denotes the previewer model. Additionally, Eq.~\ref{preview_loss} trains the previewer to follow the sampling trajectory without $\boldsymbol{c}_{txt}$, removing its dependency on text conditions which are typically unavailable in BIR tasks. The consistency constraint~\cite{song2023consistency} of enforcing consistent outputs across time-step enabling the Previewer to decode generative references on-the-fly.

\paragraph{Latent Aggregator}

The primary challenge in the BIR task is the input distribution shift. Previous work address this by aligning LQ features with reference HQ images or a learned feature codebook. The former takes extra inputs, while the latter is limited to a specific domain by the codebook capacity. In contrast, we generate reference features directly from diffusion prior. Since the compact embedding $\boldsymbol{c}_{lq}$ retains only high-level information, it is insufficient for the Previewer to reconstruct HQ images at larger time-steps, as shown in Fig.~\ref{preview_row}. Relying solely on reference preview incurs error accumulation, so the Aggregator anchors preview to the original input to prevent divergence in the reverse process. The input LQ image is encoded into SD's latent space and spatially concatenated with the preview. This expanded input remains compatible to the diffusion UNet, allowing the Aggregator to be initialized as a trainable copy of UNet compression path following~\cite{zhang2023adding}. We remove text cross-attention layers to make the Aggregator lightweight and independent of textual conditions like the Previewer. The preview and LQ hidden featrues are fused in the spatial-attention layers, which are further integrated via Spatial Feature Transform (SFT)~\cite{wang2018recovering}. For hidden feature $\boldsymbol{H}^l$ at the $l$-th layer in the Aggregator, we first split it spatially into $\boldsymbol{h}_{p}^{l}$ and $\boldsymbol{h}_{o}^{l}$, corresponding to the hidden features of preview and LQ latent, and integrate them with SFT:
\begin{equation}
\label{sft}
\boldsymbol{h}^l_{res}=\left(1+\boldsymbol{\alpha}^{l}\right)\odot \boldsymbol{h}_{p}^{l}+\boldsymbol{\beta}^{l}; \boldsymbol{h}_{p}^{l},\boldsymbol{h}_{o}^{l}=\texttt{Split}\left( \boldsymbol{H}^{l} \right),
\end{equation}
where $\boldsymbol{\alpha}^{l}, \boldsymbol{\beta}^{l}=\mathcal{M}_{\theta}^{l}(\boldsymbol{h}_{o}^l)$ are two affine transformation parameters calculated from the feature map of LQ latent at this level. We extract multi-level features $\left\{ \boldsymbol{h}_{res}^{l} \right\}_{l=1}^L$ from Aggregator using Eq.~\ref{sft}, and inject them into the corresponding part of U-Net expansion path through residual connections.

\subsection{Adaptive Restoration}

\begin{figure}[t]
    \centering
    \includegraphics[width=1.0\linewidth]{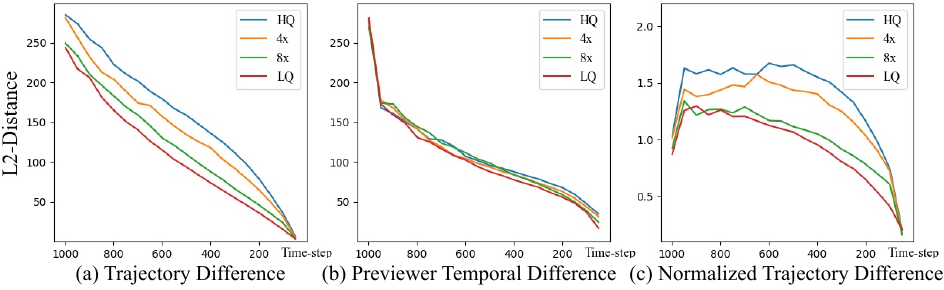}
    \caption{The evolution of the Previewer outputs during generation. (a) L2-distances between previews and denoising means; (b) temporal differences of the Previewer trajectory, measured by L2-distances between adjacent points; (c) relative distances between previews and denoising means.}
    \label{fig:trajectory}
\end{figure}


\methodname processes LQ image through two complementary ways: 1) extracting compact representation using the DCP, which is robust to degradation but loses fine-grained information; 2) encoding via the lossless SD-VAE and integrating with restoration preview, which is prone to errors in the SD-VAE. Under severe degradation, \methodname may produce samples deviate from the target HQ image. In such cases, restoration previews exhibit small variation, suggesting the DCP struggles to provide guidance according to the input. We further analyze the trajectory of restoration previews during the reverse process, compare it with the denoising predictions from Eq.~\ref{ddim}. We assess them on four degradation levels: HQ image, 4x downsampling, 8x downsampling and synthetic multi-degradation, representing decreasing input quality. Fig.~\ref{fig:trajectory} (a) illustrates the L2-distance between these two trajectories, which increases monotonically as input quality improves. A pronounced disparity between preview and ordinary denoising prediction represents the Previewer is confident with the guidance, suggesting the input LQ image is informative. Based on this observation, we use the relative difference between two predictions as an indicator of input quality:
\begin{equation}
\label{relative_dis}
\delta=\frac{\lVert \Psi \left(\boldsymbol{z}_{t},t, \Phi\left(\boldsymbol{c}_{lq}, t\right)\right) - \boldsymbol{\hat{z}}_{t}\rVert^2}{\lVert \Psi \left(\boldsymbol{z}_{t},t, \Phi\left(\boldsymbol{c}_{lq}, t\right)\right) - \Psi \left(\boldsymbol{z}_{t+1},t+1, \Phi\left(\boldsymbol{c}_{lq}, t+1\right)\right) \rVert^2},
\end{equation}
where $\boldsymbol{\hat{z}}_{t}$ is given by Eq.~\ref{ddim}. From Fig.~\ref{fig:trajectory} (b) we can see the Previewer is unstable at the beginning. The consistency training in Eq.~\ref{preview_loss} drives it to decode aggressively, causing large prediction variance during early reverse process where the input diffusion latent is too noisy. Normalizing the L2-distance between trajectories with Previewer's temporal difference effectively mitigates the temporal correlation as illustrated in Fig.~\ref{fig:trajectory} (c). A larger $\delta$ indicates higher input quality, and the conditional signals from the Aggregator should be amplified to preserve fine-grained information from the original input. On the other hand, DPM is known to first generate low-frequency features such as global structure, and add high-frequency details in the later stage of the reverse process. A decreasing $\delta$ encourages \methodname to produce diverse by exploiting the LQ image representation. We provide pseudo-code of the proposed adaptive restoration (AdaRes) algorithm in Alg.~\ref{adares}.

\begin{wrapfigure}{r}{0.5\textwidth}
\vspace{-20pt}
\begin{minipage}{0.5\textwidth}
\begin{algorithm}[H]
\caption{Adaptive Restoration}
\label{adares}
\begin{algorithmic}[1]
\Require $\boldsymbol{\epsilon_{\theta}}, \boldsymbol{\Psi}, \boldsymbol{\Phi}, \boldsymbol{c}, \alpha, \beta, \eta$
\State Sample $\boldsymbol{z}_T \sim \mathcal{N}(\textbf{0}, \beta_{T}\boldsymbol{I})$
\State Initialize $\boldsymbol{\psi} = \textbf{0}, \boldsymbol{z} = \textbf{0}, \delta = 1$
\For{$t$ in $[T, \dots, 1]$}
    \State $\boldsymbol{\hat{\psi}} = \boldsymbol{\Psi}(\boldsymbol{z}_t, t, \boldsymbol{c})$
    \State $\boldsymbol{\hat{z}} = (\boldsymbol{z}_t - \beta_t \boldsymbol{\epsilon_{\theta}}(\boldsymbol{z}_t, t, \boldsymbol{c}, \boldsymbol{\hat{\psi}}, \delta)) / \alpha_t$
    \If{$t > \eta$}
        \State $\delta = \| \boldsymbol{\hat{\psi}} - \boldsymbol{\hat{z}} \|^2 \cdot \| \boldsymbol{\hat{\psi}} - \boldsymbol{\psi} \|^{-2}$
    \Else
        \State $\delta = 0$
    \EndIf
    \State $\boldsymbol{\psi} = \boldsymbol{\hat{\psi}}, \boldsymbol{z} = \boldsymbol{\hat{z}}$
    \State $\boldsymbol{z}_{t-1} = (\beta_{t-1}/\beta_{t})\boldsymbol{z}_{t} - (\alpha_{t}/\beta_{t}-\alpha_{t-1})\boldsymbol{\hat{z}}$
\EndFor
\Ensure $\boldsymbol{z}_{0}$
\end{algorithmic}
\end{algorithm}
\end{minipage}
\vspace{-20pt}
\end{wrapfigure}

Surprisingly, although only the DCP module is explicitly trained on text-image data, \methodname demonstrates notable creativity following textual descriptions. By employing a text-guided Previewer, we can generate diverse restoration variations with compound semantics from both modalities. However, these variation samples can conflict with the original input, making them ineligible as generative references. Inspired by previous work in image editing, we disable the Aggregator at later stage generation and let \methodname renders semantic details according to LQ representation and text prompt. This ensures the low-frequency features are succeeded from the Aggregator, meanwhile prevents the high-frequency semantics and noise from entering the final results.
\section{Experiments}
\label{experiment}

\subsection{Implementation Details}

\methodname is built on SDXL~\cite{podell2023sdxl} accompanied by a two-stage training strategy. In Stage-\uppercase\expandafter{\romannumeral1}, we train the DCP module on a frozen SDXL, followed by consistency distillation of the Previewer (see Sec.~\ref{previewer}). The Previewer is trained by applying Low-Rank Adaptation (LoRA)~\cite{hu2021lora} on the base SDXL model for efficiency. By toggling the Previewer LoRA, we can seamlessly switch between the Previewer and SDXL, reducing memory footprint. After obtaining the DCP and Previewer LoRA, we proceed to Stage-\uppercase\expandafter{\romannumeral2} Aggregator training. The two-stage training ensures the Aggregator receives high-quality previews since the beginning of its training course.

We adopt SDXL's data preprocessing and conduct training on $1024^2$ resolution. In both two stages we use the AdamW~\cite{loshchilov2017decoupled} optimizer with a learning rate of $1\times 10^{-4}$. In Stage-\uppercase\expandafter{\romannumeral1}, we train the DCP module using a batch size of 256 over 200K steps, and distill the Previewer for another 30K steps with the same batch size. We train the Aggregator with a batch size of 96 over 200K steps in Stage-\uppercase\expandafter{\romannumeral2}. The entire training process spans approximately 9 days on 8 Nvidia H800 GPUs.

To enable Classifier-free Guidance (CFG)~\cite{ho2022classifier} sampling, we apply LQ image dropout with a probability of 15\% in both stages training. In all test experiments, we employ 30 steps DDIM sampling~\cite{song2020denoising} with a CFG scale of $7.0$.

\begin{table}[t]
\centering
\caption{Quantitative comparisons on both synthetic validation data and public real-world dataset. We highlight the best results in \textbf{bold} and the second best with \underline{underline}.}
\label{tab:quantitative}
\begin{subtable}[h]{\textwidth}
\resizebox{1.0\textwidth}{!}{
    \begin{tabular}{llcccccc}
    \toprule
    \multicolumn{1}{c}{\bf Dataset} & \multicolumn{1}{c}{\bf Model} & \bf PSNR & \bf SSIM & \bf LPIPS & \bf CLIPIQA & \bf MANIQA & \bf MUSIQ \\
    \midrule
    \multirow{6}{*}{Synthetic}    &BSRGAN  &20.21   &0.5214 &0.7793  &0.2072  &0.2076  &17.53\\
    &Real-ESRGAN   &19.92  &\underline{0.5317} &0.7554   &0.2102  &0.2331  &17.39\\
    &StableSR   &\underline{20.42}  &\textbf{0.5388} &\underline{0.3751}   &0.4672  &0.2602  &52.33\\
    &CoSeR   &19.92  &0.5114 &\textbf{0.3353}   &\textbf{0.6651}  &\underline{0.4152}  &\underline{67.51}\\
    &SUPIR   &\textbf{20.46}  &0.4990 &0.4090   &0.4875  &0.3081  &56.43\\
    &\methodname (ours)   &18.54  &0.5126 &0.3986   &\underline{0.5497}  &\textbf{0.4379}  &\textbf{68.59}\\
    \midrule
    \multirow{6}{*}{Real-world}    &BSRGAN  &26.38   &0.7651 &0.4120  &0.3151  &0.2147  &28.58\\
    &Real-ESRGAN   &\textbf{27.29}  &\textbf{0.7894} &0.4173   &0.2532  &0.2398  &25.66\\
    &StableSR   &26.40  &\underline{0.7721} &\textbf{0.2597}   &0.4501  &0.2947  &48.79\\
    &CoSeR   &25.59  &0.7402 &\underline{0.2788}   &\textbf{0.5809}  &\underline{0.3941}  &\underline{60.51}\\
    &SUPIR   &\underline{26.41}  &0.7358 &0.3639   &0.3869  &0.2721  &42.72\\
    &\methodname (ours)   &21.75  &0.6766 &0.3686   &\underline{0.5401}  &\textbf{0.4819}  &\textbf{65.32}\\
    \bottomrule
    \end{tabular}
}
\label{tab:scheme1}
\caption{Scenario 1: $512^2$ image restoration. The outputs of SUPIR and \methodname are downsampled to $512^2$.}
\end{subtable}
\begin{subtable}[h]{\textwidth}
\resizebox{1.0\textwidth}{!}{
    \begin{tabular}{llcccccc}
    \toprule
    \multicolumn{1}{c}{\bf Dataset} & \multicolumn{1}{c}{\bf Model} & \bf PSNR & \bf SSIM & \bf LPIPS & \bf CLIPIQA & \bf MANIQA & \bf MUSIQ \\
    \midrule
    \multirow{6}{*}{Synthetic}    &BSRGAN  &\textbf{21.32}   &\underline{0.5267} &0.5611  &0.4289  &0.3299  &37.97\\
    &Real-ESRGAN   &20.45  &0.5202 &0.5660   &0.4566  &0.3627  &37.92\\
    &StableSR   &\underline{21.01}  &\textbf{0.5490} &0.3921   &0.4526  &0.2492  &48.94\\
    &CoSeR   &20.50  &0.5215 &\textbf{0.3488}   &\textbf{0.6461}  &0.3939  &\underline{64.84}\\
    &SUPIR   &20.57  &0.4569 &0.4196   &\underline{0.6286}  &\underline{0.3962}  &61.00\\
    &\methodname (Ours)   &18.80  &0.5076 &\underline{0.3903}   &0.6111  &\textbf{0.4303}  &\textbf{66.09}\\
    \midrule
    \multirow{6}{*}{Real-world}    &BSRGAN  &\textbf{28.60}   &\underline{0.8141} &0.3690  &0.4720  &0.2258  &18.26\\
    &Real-ESRGAN   &\underline{28.13}  &\textbf{0.8209} &0.3647   &0.4435  &0.3229  &35.31\\
    &StableSR   &27.79  &0.8043 &\textbf{0.2514}   &0.4634  &0.2901  &46.54\\
    &CoSeR   &27.04  &0.7683 &\underline{0.2882}   &\textbf{0.5847}  &\underline{0.4068}  &\underline{58.39}\\
    &SUPIR   &26.10  &0.5825 &0.5429   &0.4822  &0.3232  &44.95\\
    &\methodname (Ours)   &21.89  &0.6879 &0.3601   &\underline{0.5647}  &\textbf{0.4389}  &\textbf{62.58}\\
    \bottomrule
    \end{tabular}
}
\label{tab:scheme2}
\caption{Scenario 2: $1024^2$ image restoration. We crop $512^2$ patches as inputs to 512-models and evaluate the quantitative metrics on the cropped area only.}
\end{subtable}
\end{table}

\subsection{Experimental Configuration}

\paragraph{Training Data}

We synthesis LQ-HQ image pairs using Real-ESRGAN~\cite{wang2021real} with the default setting. As mentioned in Sec.~\ref{previewer}, we conduct Stage-\uppercase\expandafter{\romannumeral1} training on the JourneyDB dataset~\cite{sun2024journeydb}, a generated dataset with descriptive captions. While JourneyDB images are of extreme quality, they lack the textures in real-world images. Hence for Stage-\uppercase\expandafter{\romannumeral2} training, we incorporate publicly available texture-rich datasets to enhance model's ability to produce realistic visual details. Specifically, we use DIV2K~\cite{agustsson2017ntire}, LSDIR~\cite{li2023lsdir}, Flickr2K~\cite{timofte2017ntire} and FFHQ~\cite{karras2019style}.

\paragraph{Test Setting}

For a comprehensive evaluation, we test \methodname on a synthetic dataset and public benchmarks following previous work. We synthesize $2,000$ multi-degradation samples from DIV2K and LSDIR validation sets using Real-ESRGAN pipeline, filtering out images smaller than $1024^2$ to ensure ground-truth quality. We include a small portion of JourneyDB validation data to enhance benchmark diversity. We conduct evaluations on RealSR~\cite{cai2019toward} and DRealSR~\cite{wei2020component} to assess model performance on real-world LQ images. We report full-reference metrics PSNR, SSIM, LPIPS~\cite{zhang2018unreasonable}, if ground-truth targets are available, and non-reference metrics MANIQA~\cite{yang2022maniqa}, CLIPIQA~\cite{wang2023exploring}, MUSIQ~\cite{ke2021musiq} to quantitatively compare \methodname with other models.

\subsection{Comparing to Existing Methods}

We compare \methodname with state-of-the-art models, including StableSR~\cite{wang2024exploiting}, CoSeR~\cite{sun2024coser}, SUPIR~\cite{yu2024scaling}, BSRGAN~\cite{zhang2021designing} and Real-ESRGAN~\cite{wang2021real}. Since some of them are limited to $512^2$ resolution, we consider two test scenarios for a fair comparison: 1) models are tested on $512^2$ images with outputs of 1024-models scaled accordingly; 2) following SUPIR, the models are tested on $1024^2$ images by cropping $512^2$ patch as inputs to 512-models, metrics are evaluated on the cropped area only.

\paragraph{Quantitative Comparison}

\begin{figure}[t]
\centering
    \includegraphics[width=\linewidth]{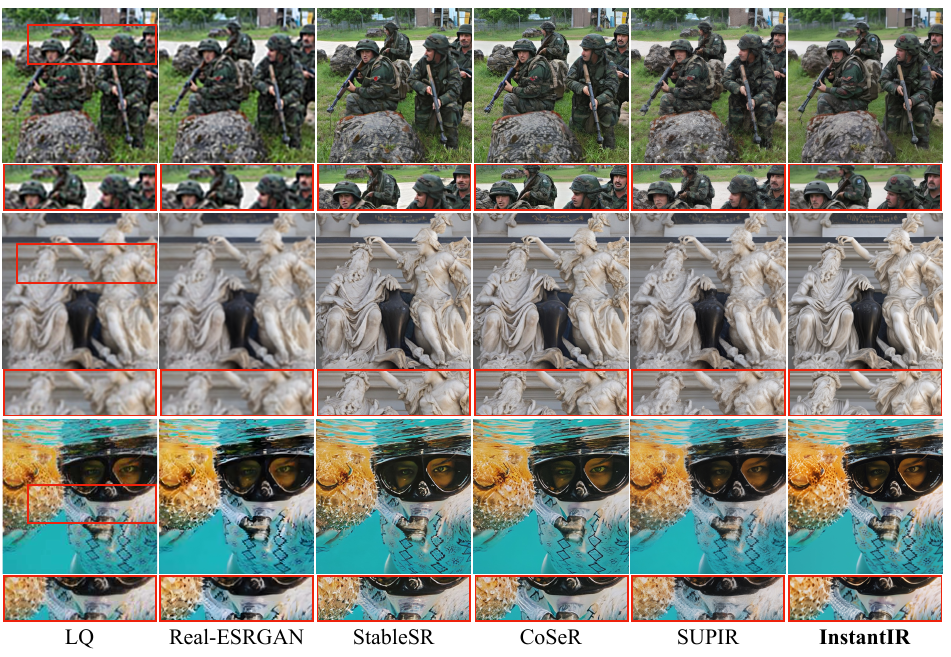}
    \caption{Qualitative comparisons on real-world LQ images. Restorations from \methodname are rich in details with global semantic consistency. Better viewed zoom in.}
    \label{qualitative_real}
\end{figure}

The results are summarized in Tab.~\ref{tab:quantitative}. \methodname continuously achieves the highest MUSIQ and MANIQA scores across all test settings, outperfoming the second best by large margins up to \textbf{22\%} in MANIQA and \textbf{8\%} in MUSIQ. Notably in scenario 1, despite halving the input data, \methodname still performs comparably to SOTA models. While CoSeR achieves the best CLIPIQA scores closely followed by \methodname, restorations from 1024-models SUPIR and \methodname are rich in details as shown in Fig.~\ref{qualitative_real}. We also observe the misalignment of PSNR and SSIM scores with visual quality as reported in the literature~\cite{yu2024scaling,wang2024exploiting}. We include these metrics here for reference purpose.

\paragraph{Qualitative Comparison}

We provide some restoration samples on real-world LQ images in Fig.~\ref{qualitative_real}. Through leveraging the previewing mechanism, \methodname actively aligns with generative prior, reducing hallucinations and producing sharp yet realistic details. In the second row of Fig.~\ref{qualitative_real}, while SUPIR's result contains rich textures, the absence of global semantic guidance causes the diver’s body and mask to blend together. In contrast, the cognitive encoder in CoSeR helps it identifies statues in the second example. CoSeR employs a feature codebook to handle unknown degradations, which limits the generation of complex textures on the statues. Notably in the first row of Fig.~\ref{qualitative_real}, \methodname is the only one that successfully recovers all four faces without distortion, suggesting its superior ability in capturing semantic and reproduce realistic details from diverse degradations.

\begin{figure}[t]
\centering
\begin{subfigure}{0.49\textwidth}
    \includegraphics[width=\linewidth]{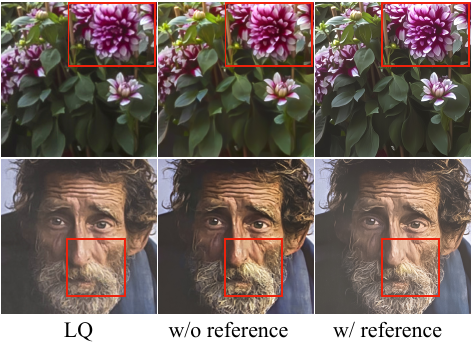}
    \caption{In-domain previews enhance detail restoration.}
    \label{indomain_preview}
\end{subfigure}
\begin{subfigure}{0.49\textwidth}
    \includegraphics[width=\linewidth]{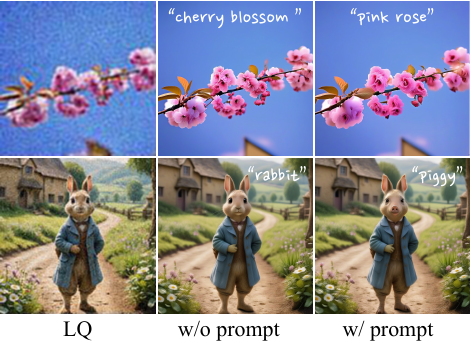}
    \caption{Out-domain previews edits high-level semantics.}
    \label{outdomain_edit}
\end{subfigure}
\caption{Visual examples of the previewing mechanism in \methodname. Better viewed zoom in.}
\end{figure}

\begin{figure}[t]
\centering
    \includegraphics[width=\linewidth]{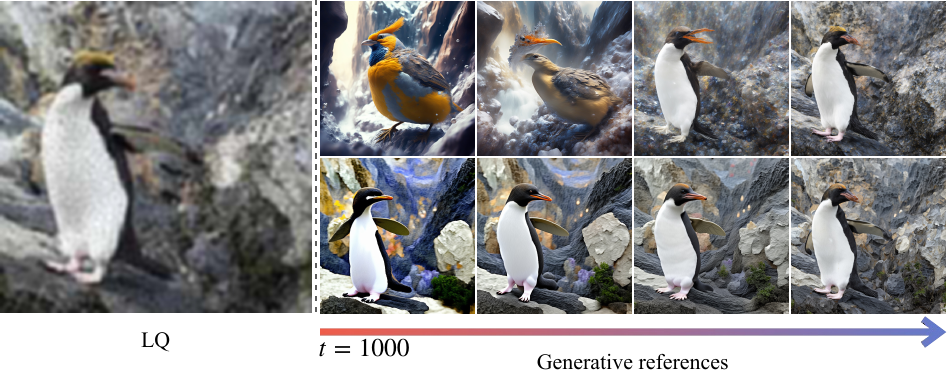}
    \caption{Visual examples of generative references. The first row on the right is generated by DCP trained on image-only data, while the second row is produced by DCP used in \methodname, which is trained on text-image pairs.}
    \label{preview_row}
\end{figure}

\subsection{Restoration with Previewing}

\paragraph{In-domain Reference for Detail Enhancement}

Reference-based BIR models improve detail restoration by transferring high-quality textures from HQ references. \methodname achieves this by querying the T2I model, eliminating additional inputs. In Fig.~\ref{indomain_preview}, we disable the Previewer to see the effect of generative references. Here \methodname infers solely with LQ images, which is beneficial to fidelity preservation but bad for visual quality. This is also reflected in Tab.~\ref{ablation2} where all quantitative metrics deteriorate except PSNR and SSIM. Moreover, \methodname equipped with Alg.~\ref{adares} further improves the non-reference metrics, suggesting its flexibility in to different conditions.

\paragraph{Out-domain Reference for Creative Restoration}

Fig.~\ref{outdomain_edit} shows more creative restoration samples. Owing to the efficiency of our Aggregator in integrating reference latents, \methodname is able to perform high-level semantic editing during restoration, altering specific attributes of the subject and leaving other visual details like global structure and layout intact. We empirically find \methodname offer better prompt-following ability under heavy degradation.

\begin{table}[t]
\centering
\caption{Ablation studies. The best results are highlighted in \textbf{bold}.}
\begin{subtable}[h]{\textwidth}
\centering
    \begin{tabular}{lcccccc}
    \toprule
    & \bf PSNR & \bf SSIM & \bf LPIPS & \bf CLIPIQA & \bf MANIQA & \bf MUSIQ \\
    \midrule
    Baseline  &21.40    &0.6775  &\textbf{0.3173}  &\textbf{0.5433}  &\textbf{0.4024}  &\textbf{66.35}\\
    +Distillation  &\textbf{24.24}    &0.6963  &0.4306  &0.2453  &0.2145  &38.33\\
    +Noisy Previews  &23.07    &\textbf{0.7312}  &0.3830  &0.3767  &0.2924  &49.23\\
    \bottomrule
    \end{tabular}
\caption{Ablation study of the consistency distillation in Previewer and adding fresh noise to restoration previews.}
\label{ablation1}
\end{subtable}
\begin{subtable}[h]{\textwidth}
\centering
\resizebox{1.0\textwidth}{!}{
    \begin{tabular}{cccccccc}
    \toprule
    \bf References & \bf AdaRes & \bf PSNR & \bf SSIM & \bf LPIPS & \bf CLIPIQA & \bf MANIQA & \bf MUSIQ \\
    \midrule
    \xmark & \xmark  &\textbf{22.24}    &\textbf{0.7539}  &0.3672  &0.2721  &0.2128  &42.64\\
    \cmark & \xmark  &21.13    &0.6728  &\textbf{0.3173}  &0.5445  &0.3747  &64.86\\
    \cmark & \cmark  &21.06    &0.6708  &0.3189  &\textbf{0.5456}  &\textbf{0.3766}  &\textbf{64.94}\\
    \bottomrule
    \end{tabular}
}
\caption{Ablation study of the generative references and AdaRes sampling.}
\label{ablation2}
\end{subtable}
\end{table}

\subsection{Ablation Study}

\paragraph{DCP Training on Text Domain}

We compare training DCP module with and without textual condition. Due to limited computational resources, we did not proceed to train the subsequent Previewer and Aggregator for the DCP trained on image-only data. For comparison, we provide some visual examples of their generative references across diffusion time-steps in Fig.~\ref{preview_row}. As shown in the first row of Fig.~\ref{preview_row}, the generative references from the image-only DCP differ significantly from the input LQ image at early stage, retaining only coarse semantic like ``a bird standing on a rocky surface.'' In contrast, DCP trained with text descriptions preserves most of the low-level information, including global hue, structure, layout, and even the subject's category (penguin) and its pose.

\paragraph{Previewer Consistency Distillation}

We validate the necessity of consistency constraints in Previewer. We experiment with using predictions from Eq.~\ref{ddim} as reference inputs to the Aggregator. The second row in Tab.~\ref{ablation2} shows a significant drop in the non-reference metrics. In fact, the prediction in Eq.~\ref{ddim} is close to the distribution mean at each time-step~\cite{karras2022elucidating}. Previewer with consistency distillation can directly sample from the data distribution, providing more informative generative references.

\paragraph{Fresh Noise to Restoration Previews}

We additionally train an Aggregator that injects fresh noise to reference latents according to diffusion time-step. The noisy preview latent follows the same distribution as current diffusion latent, making the overall pipeline resemble a ControlNet model~\cite{zhang2023adding}. As shown in the third row of Tab.~\ref{ablation2}, \methodname significantly outperforms ControlNet with LQ image as conditional inputs. This highlights the effectiveness of the previewing mechanism in \methodname for adjusting generation conditions during inference.
\section{Conclusion}

In this paper, we explore a novel method to address unknown degradations in BIR tasks. Through exploiting the generation process of DPM, we propose to actively align with the generative prior to reduce the errors in encoding LQ image. Our pipeline is implemented based on pre-trained SDXL model, referred to as \methodname. Extensive experiments demonstrate the exceptional restoration capability of \methodname, delivering SOTA performance in quantitative metrics and visual quality. However, we observe some disparity in reference metrics such as PSNR and SSIM compared to SOTA models, which might because of the excessive generative prior diminishes fidelity. Future work will explore approach to improve the interaction between generative prior and conditions, as well as ways to refine the previewer to produce more reliable references.


\bibliography{reference}
\bibliographystyle{iclr2025_conference}

\end{document}